\documentclass{article}
\usepackage{spconf,amsmath,epsfig}
\usepackage{amssymb}
\usepackage{mathrsfs}
\usepackage{epstopdf}
\usepackage{algorithmic}
\usepackage{algorithm}
\usepackage{subfigure}

\title{MAP-GUIDED HYPERSPECTRAL IMAGE SUPERPIXEL SEGMENTATION USING PROPORTION MAPS}
\name{Hao Sun\textsuperscript{*} and Alina Zare\textsuperscript{$\dagger$}
\thanks{The authors wish to thank the National Geospatial-Intelligence Agency for support of this research under the project entitled ``NIP: Functions of Multiple Instances for Hyperspectral Analysis.''}
}
\address{Department of Electrical and Computer Engineering, University of Missouri\textsuperscript{*} \\
Department of Electrical and Computer Engineering, University of Florida\textsuperscript{$\dagger$}}
\begin{document}
%
\maketitle
\begin{abstract}
A map-guided superpixel segmentation method for hyperspectral imagery is developed and introduced. The proposed approach develops a hyperspectral-appropriate version of the SLIC superpixel segmentation algorithm, leverages map information to guide segmentation, and incorporates the semi-supervised Partial Membership Latent Dirichlet Allocation (sPM-LDA) to obtain a final superpixel segmentation. The proposed method is applied to two real hyperspectral data sets and quantitative cluster validity metrics indicate that the proposed approach outperforms existing hyperspectral superpixel segmentation methods. 
\end{abstract}
\section{Introduction}
\label{sec:intro}

%
%

Many effective superpixel image segmentation algorithms have been developed in the literature for gray-scale or RGB imagery. However, very few of these approaches are directly applicable to hyperspectral imagery. 
In \cite{sethi2015scalable}, the ultrametric contour map (UCM) algorithm was extended to hyperspectral imagery (HSI) through the use of principal component analysis to reduce the image dimensionality to three dimension such that UCM can be directly applied. The normalized cuts algorithm has also been extended for hyperspectral imagery \cite{gillis2012hyperspectral}.  In \cite{thompson2010superpixel}, a hyperspectral superpixel algorithm was developed using a graph-based approach that relied on the sum of squared differences between neighboring pixels.  


The superpixel segmentation approach proposed here differs from existing hyperspectral superpixel segmentation methods primarily in two ways: (1) the proposed method uses proportion maps obtained through unmixing as a dimensionality-reduced version of the image scene; and (2) the method relies on map data (specifically, data from crowdsourced OpenStreetMap  \cite{haklay2008openstreetmap}) to guide the unmixing and segmentation.

\section{Method}
\label{sec:pagestyle}
The proposed map-guided hyperspectral unmixing-based superpixel segmentation algorithm has three stages.  Stage one consists of obtaining an initial superpixel segmentation result using a modified version of SLIC \cite{achanta2012slic}.  Stage two consists of applying the semi-supervised Partial Membership Latent Dirichlet Allocation (sPM-LDA) to obtain unmixing results \cite{zou2016hyperspectral}. Finally, stage three obtains a final superpixel segmentation by clustering the proportion vectors from stage two.  

\subsection{Stage 1: Map-Guided Hyperspectral SLIC}
\label{sec:stage1}
To perform an initial superpixel segmentation, we extend Simple Linear Iterative Clustering (SLIC) \cite{achanta2012slic} to hyperspectral imagery. Previously in the literature, dimensionality reduction was applied to HSI prior to allow for application of SLIC \cite{zhang2015slic}. In this paper, we develop a new approach for extending SLIC to HSI. In our method, no dimensionality reduction is applied.  Instead, the full spectral information is used (as opposed to the $Lab$ features used in SLIC originally) in combination with spatial information. The proposed hyperspectral SLIC (HSLIC) is iterative: (1) initial cluster centers are obtained with a regular spatial sampling of the image where the cluster centers are vectors in which spectral information is concatenated with spatial coordinates; (2) then, each pixel is assigned to a nearest cluster center using a distance measure that includes a spectral and a spatial term; and (3) cluster centers are then updated to be equal to the average of all pixels assigned to the cluster. This process is iterated until the convergence. 

\begin{figure}[h!]
  \centering
    \includegraphics[width=0.3\textwidth]{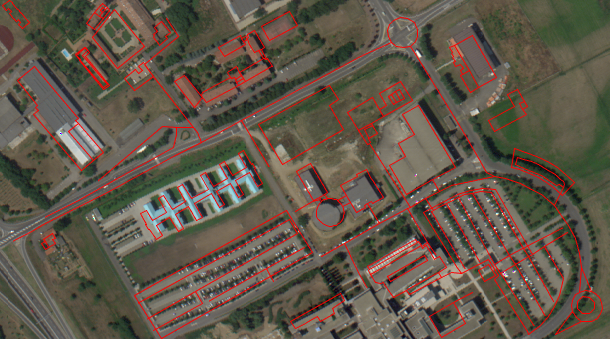}
  \caption{OpenStreetMap polygons overlaid on the Pavia University HSI image.}
  \label{fig:OSM_geo}
\end{figure}

To leverage map information, our map-guided SLIC algorithm merges superpixels that intersect with a common map element.  For example, from OpenStreetMap (OSM), map elements such as roads and building profiles can be obtained, as shown in Fig. \ref{fig:OSM_geo}. These map elements can be roughly aligned to the hyperspectral imagery using a affine transformation. In our implementation, this transformation is obtained using manual selection of corresponding points. However, approaches such as map conflation can be used as well \cite{song2009automated}. After alignment, the superpixels obtained using hyperspectral SLIC method described above which overlap with shared map polygons are merged into one superpixel. The full map-guided hyperspectral SLIC is summarized in Algorithm \ref{alg:SLIC}.

\begin{algorithm}[ht]
\begin{algorithmic}[1]
    \REQUIRE HSI Data, $K$ (number of superpixels), $m$ (scaling factor)
\STATE Initialize cluster centers $\left\{\mathbf{c}_k\right\}_{k=1}^K$ by sampling pixels at regular grid steps.
\STATE Perturb  $\left\{\mathbf{c}_k\right\}_{k=1}^K$ in an $n \times n$ neighborhood to the lowest gradient position.
\STATE \textbf{Repeat}
\FOR {each  $\mathbf{c}_k$}
\STATE Calculate the spectral distance of pixel $\mathbf{x}_i$ and $\mathbf{c}_k$ over all bands $\lambda$: $d_{spectral} = \sum^B_{\lambda=1} \| \mathbf{x}_i(\lambda) -  \mathbf{c}_k(\lambda)\|^2_2$
\STATE Calculate the spatial distance of pixel $\mathbf{x}_i$ and $\mathbf{c}_k$:\\ $d_{spatial} = \sqrt{(a_{\mathbf{x}_i}-a_{\mathbf{c}_k})^2+(b_{\mathbf{x}_i}-b_{\mathbf{c}_k})^2}$ where $a,b$ are pixel coordinates.
\STATE   Assign each pixel to a $\mathbf{c}_k$ in a $2S \times 2S$ square neighborhood based on the minimum spectral and spatial distance: $d_{\mathbf{x}_i,\mathbf{c}_k} = d_{spectral} + \frac{m}{S} d_{spatial}$
\ENDFOR
\STATE Update cluster centers as the mean of all pixels assigned to the cluster.
\STATE \textbf{ Until} stopping criterion is reached.
\STATE Align map data to imagery
\FOR {Each Map Polygon}
\STATE Merge all superpixels that overlap with this polygon
\ENDFOR
\end{algorithmic}
\caption{Map-Guided Hyperspectral SLIC Superpixel Segmentation}
\label{alg:SLIC}
\end{algorithm}

\subsection{Stage 2: Unmixing with Semi-supervised PM-LDA}
After obtaining the map-guided SLIC superpixels, the hyperspectral is unmixed to obtain proportion maps.  These proportion maps are used as a dimensionality reduced version of the imagery which are then used to obtain the final superpixel segmentation.  In this work, we use the semi-supervised partial membership latent Dirichlet allocation (sPM-LDA) to unmix the input imagery and obtain proportion maps. sPM-LDA is described in \cite{zou2016hyperspectral}. sPM-LDA uses an input superpixel segmentation to guide unmixing. Also, sPM-LDA can leverage partial label information to improve unmixing results. In this work, we use the polygon labels obtained from OSM to provide partial labels to the sPM-LDA approach. For example, superpixels obtained through merging in stage 1 that overlap with building polygons are given a partial label of ``building.'' 

%

\subsection{Stage 3: Final Superpixel Segmentation}
\label{sec:stage3}
$K$-means clustering is applied to the proportion vectors that are obtained in Stage 1.  Then, connected components analysis is applied to relabel each spatial cluster as an individual superpixel.  Finally, a ``clean-up'' post-processing is done to merge small disjoint segments into the largest neighboring superpixel. Specifically, if the number of pixels in a superpixel is less than a prescribed threshold, those pixels are merged into the largest neighboring superpixel. 

 


\section{Experimental Results}
\label{sec:typestyle}
Our proposed superpixel segmentation method is applied to two real hyperspectral data sets. The first data set was collected by the Reflective Optics System Imaging Spectrometer (ROSIS) at University of Pavia, Italy in July 2002. The image contains $340 \times 610$ pixels and consists of 103 bands \cite{holzwarth2003hysens}. 

\begin{figure}[h]
\begin{center}
\subfigure[]{\includegraphics[height=4cm]{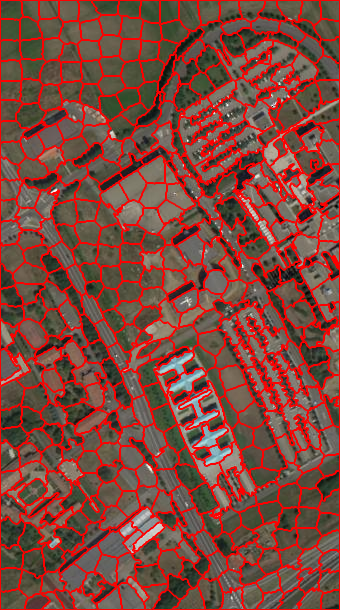}}
\subfigure[]{\includegraphics[height=4cm]{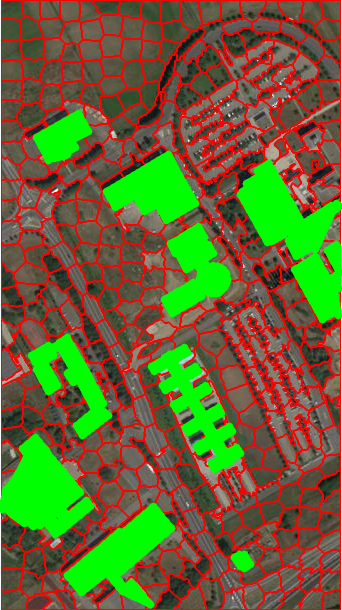}}
\subfigure[]{\includegraphics[height=4cm]{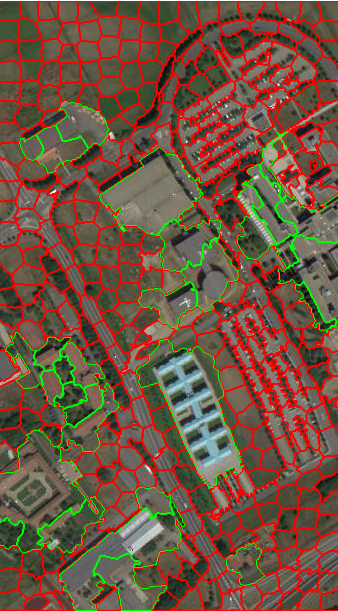}}
\caption{Superpixels on Pavia University: (a)Superpixels from SLIC; (b)Buildings extracted from OpenStreetMap; (c)Superpixels by map-guided hyperspectral SLIC}
\label{fig:pavia_initial_superpixels}
\end{center}
\end{figure}

\begin{figure}[h]
\begin{center}
\subfigure[Painted metal sheets]{\includegraphics[height=4cm]{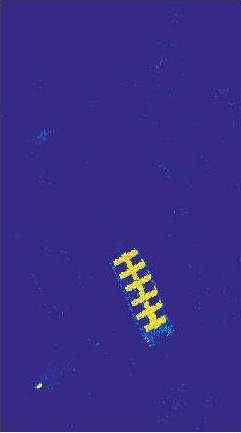}}
\subfigure[Red roof]{\includegraphics[height=4cm]{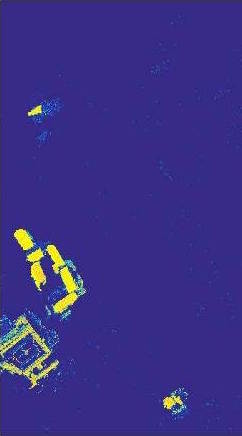}}
\subfigure[Bare soil]{\includegraphics[height=4cm]{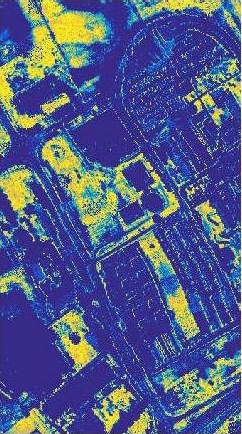}}
\subfigure[Asphalt]{\includegraphics[height=4cm]{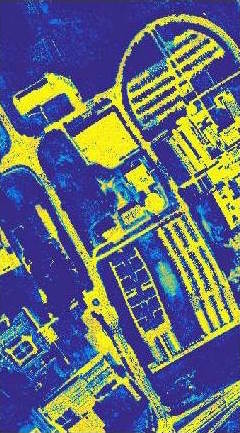}}
\subfigure[Shadow]{\includegraphics[height=4cm]{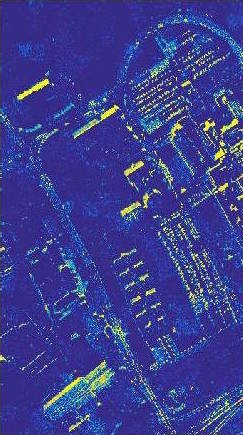}}
\subfigure[Vegetation]{\includegraphics[height=4cm]{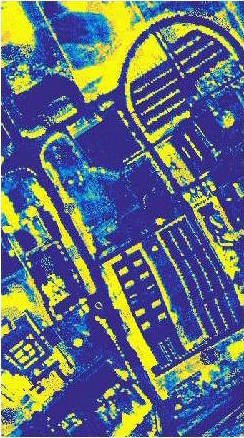}}
\caption{Estimated proportion maps using semi-supervised PM-LDA on Pavia University}
\label{fig:p_suppmlda}
\end{center}
\end{figure}

During the hyperspectral SLIC application, $K=500$ and $m=20$. During unmixing using sPM-LDA, the number of endmembers was set to $6$, $\lambda = 1$, $\alpha = 0.3$, $\epsilon = 5\%$ and $T = 200$ and the blue roof building and red roof building were partially labeled using the semi-supervised approach outlined in \cite{sheng2016unmixing}. During the final stage, $K=6$ was used during $K$-means clustering. Fig. \ref{fig:pavia_initial_superpixels} shows the results returned from hyperspectral SLIC, the extracted data from OSM and the final segmentation result. Unmixing results are shown in Fig. \ref{fig:p_suppmlda}.  

\begin{figure}[h]
\begin{center}
\subfigure[HSLIC]{\includegraphics[height=4cm]{Figures/pavia_SLIC.png}}
\subfigure[HSLIC+OSM]{\includegraphics[height=4cm]{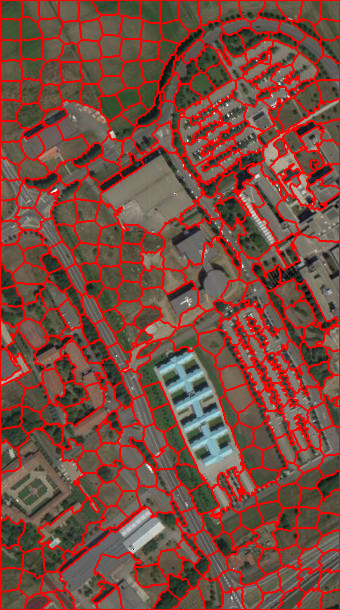}}
\subfigure[HNC]{\includegraphics[height=4cm]{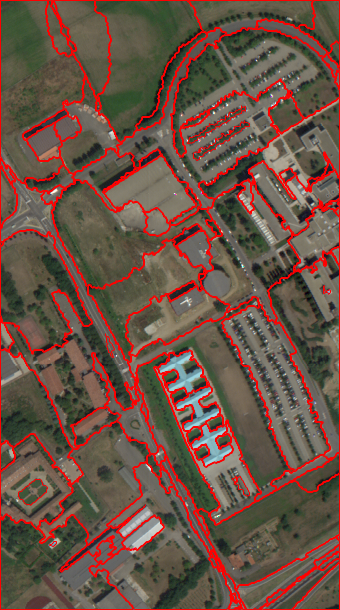}}
\subfigure[HUCM]{\includegraphics[height=4cm]{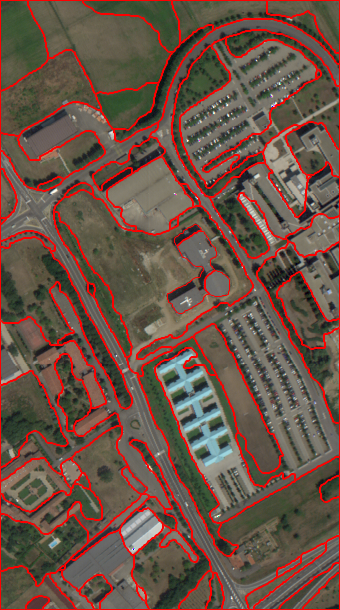}}
\subfigure[PM-LDA]{\includegraphics[height=4cm]{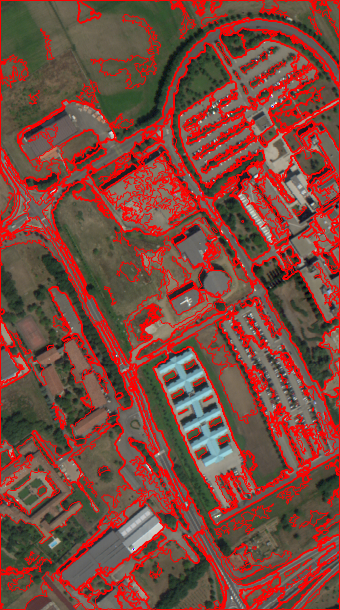}}
\subfigure[sPM-LDA]{\includegraphics[height=4cm]{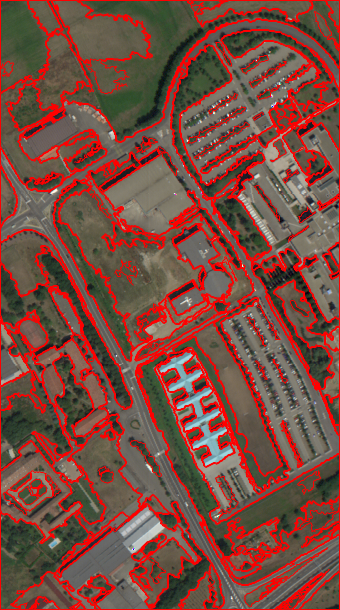}}
\caption{Superpixel Segmentation results on Pavia University}
\label{fig:pavia_superpixels}
\end{center}
\end{figure}

The proposed superpixel segmentation results as shown in Fig. \ref{fig:pavia_superpixels} are compared with four other methods: hyperspectral SLIC (HSLIC) (Sec.\ref{sec:stage1}), map-guided hyperspectral SLIC (HSLIC+OSM) (Sec.\ref{sec:stage1}), hyperspectral normalized cuts (HNC) \cite{gillis2012hyperspectral}, and hyperspectal ultra contour map (HUCM)\cite{sethi2015scalable} and a PM-LDA based superpixel segmentation (which is the same as the proposed method except without final post-processing as described in Sec.\ref{sec:stage3}).  

Examining Fig. \ref{fig:pavia_superpixels}, it can be seen that hyperspectral SLIC and hyperspectral SLIC+OSM both return oversegmented results. Results from HNC and HUCM do not oversegment but often cross distinct boundaries in the imagery resulting in incorrect superpixel segmentation. The PM-LDA based superpixel segmentation (without postprocessing) includes many very small superpixels (so small that they appear like noise). In comparison, our proposed approach can segment the imagery into semantically meaningful superpixels with regions associated with map polygons remaining intact.

To quantitatively evaluate the superpixel segmentation results, the Dunn \cite{dunn1974well}, Davies-Bouldin (DB) \cite{davies1979cluster}, Silhouette indices \cite{rousseeuw1987silhouettes} are used to measure the performance. Cluster validity measures are quantitative methods to evaluate clustering results based on intra-cluster compactness and inter-cluster separability. In our evaluation, each superpixel was treated as a distinct cluster.  Thus, good cluster validity measure values indicate that the spectral signatures within a superpixel are similar to each but distinct from other superpixels. The bigger Dunn, Silhouette indices and the smaller DB index will indicate a better partition of the image. We run each algorithm for 10 times, and calculate the means and standard deviations for these three indices. As shown in Table \ref{tab:original_indices}, the proposed method outperforms the comparison approaches. 
\begin{table}[!htb]
\centering  \
\resizebox{\columnwidth}{!}{
\begin{tabular}{|l| c| c| c |} 
\hline
 & Dunn & Davies-Bouldin & Silhouette\\  \hline     
HSLIC &$0.033\pm 0.01$ &$16.569\pm 0.61$&$-0.836\pm 0.03$\\  \hline
HSLIC+OSM &$0.040\pm 0.00$&$16.119 \pm 0.79$ &$-0.795\pm 0.03$  \\  \hline
HNC &$0.049\pm 0.02$ &$11.445 \pm 0.66$&$-0.724\pm 0.05$\\ \hline
HUCM &$0.038\pm 0.01$  &$19.296\pm 0.94$&$-0.750\pm 0.03$ \\ \hline
PM-LDA &$0.061\pm 0.01$ &$10.995\pm 0.83$&$-0.7056\pm 0.05$ \\ \hline
Proposed method &$\textbf{0.095}\pm \textbf{0.01}$&$\textbf{8.712}\pm \textbf{0.73}$ &$\textbf{-0.561}\pm \textbf{0.05}$ \\ \hline
\end{tabular}}
\caption{Dunn, Davies-Bouldin, Silhouette index results $\pm$ standard deviation on Pavia University. }
\label{tab:original_indices}
\end{table}

The proposed approach was also applied to the MUUFL Gulfport hyperspectral data \cite{gader2013muufl}.  This image was collected by the CASI-1500 hyperspectral imager over the campus of the University of Southern Mississippi-Gulfport in Long Beach, Mississippi in November 2010. The image is consisting of 72 bands with the wavelength range of 375 to 1050 nm.

\begin{figure}[h]
\begin{center}
\subfigure[HSLIC]{\includegraphics[height=3.5cm]{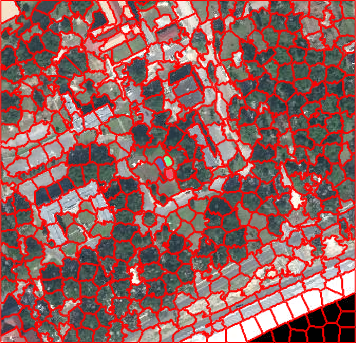}}
\subfigure[HSLIC+OSM]{\includegraphics[height=3.5cm]{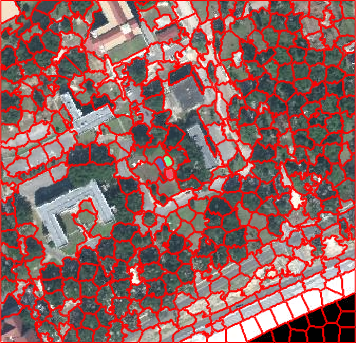}}
\subfigure[HNC]{\includegraphics[height=3.5cm]{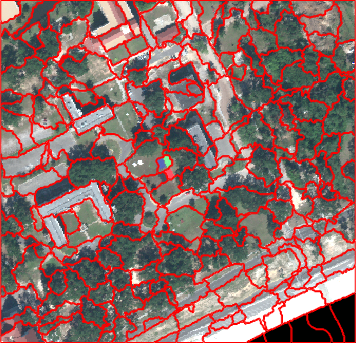}}
\subfigure[HUCM]{\includegraphics[height=3.5cm]{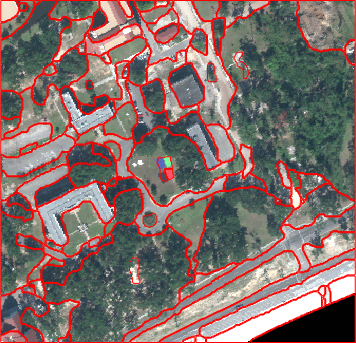}}
\subfigure[PM-LDA]{\includegraphics[height=3.5cm]{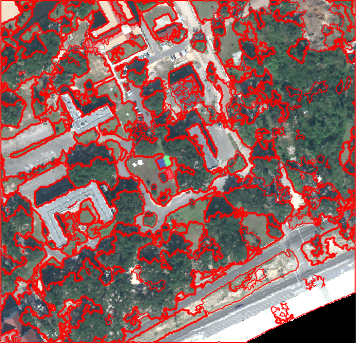}}
\subfigure[sPM-LDA]{\includegraphics[height=3.5cm]{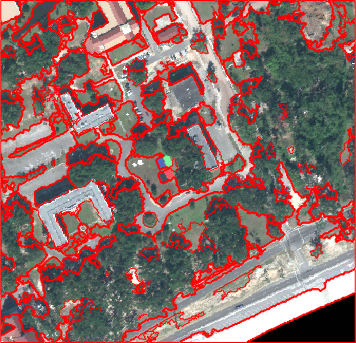}}
\caption{Superpixel Segmentation on Gulfport with 6 algorithms}
\label{fig:gulfport_superpixels}
\end{center}
\end{figure}

During the hyperspectral SLIC application, $K=500$ and $m=20$. During unmixing using sPM-LDA, the number of endmembers was set to $7$, $\lambda = 1$, $\alpha = 0.3$, $\epsilon = 10\%$ and $T = 200$ and the grey roof buildings were partially labeled using the semi-supervised approach outlined in \cite{sheng2016unmixing}. During the final stage, $K=7$ was used during $K$-means clustering. The superpixel segmentation results for the proposed method and comparison algorithms are shown in Fig. \ref{fig:gulfport_superpixels}.

The results obtained over the MUUFL Gulfport data are qualitatively similar to those obtained for Pavia University.  Quantitative evaluation shown in Table \ref{tab:original_indices_gulfport} using cluster validity metrics on MUUFL Gulfport indicates that proposed method outperforms comparison methods on this data set as well. 

\begin{table}[!htb]
\centering  \
\resizebox{\columnwidth}{!}{
\begin{tabular}{|l| c| c| c |} 
\hline
 & Dunn & Davies-Bouldin & Silhouette\\  \hline     
HSLIC &$8.220\times 10^{-4}\pm 0.00$ &$12.772\pm 0.08$&$-0.906\pm 0.03$\\  \hline
HSLIC+OSM &$9.042\times 10^{-4}\pm 0.00$&$12.638 \pm 0.08$ &$-0.882\pm 0.02$  \\  \hline
HNC &$0.014\pm 0.01$ &$12.570 \pm 0.09$&$-0.795\pm 0.03$\\ \hline
HUCM &$0.040\pm 0.01$  &$9.390\pm 0.04$&$-0.659\pm 0.03$ \\ \hline
PM-LDA &$0.067\pm 0.01$ &$8.049\pm 0.05$&$-0.546\pm 0.03$ \\ \hline
Proposed method &$\textbf{0.078} \pm \textbf{0.02}$&$\textbf{8.035}\pm \textbf{0.05}$ &$\textbf{-0.515}\pm \textbf{0.03}$ \\ \hline
\end{tabular}}
\caption{Dunn, Davies-Bouldin, Silhouette indices $\pm$ standard deviation on Gulfport}
\label{tab:original_indices_gulfport}
\end{table}



\section{Conclusion}
\label{sec:majhead}
A new map-guided unmixing-based hyperspectral image superpixel segmentation method is proposed. The proposed method obtains superpixel segmentation results that leverage any available map-information and produces a superpixel segmentation with semantically meaningful segments.  

{\small
\bibliographystyle{IEEEbib}
\bibliography{strings,refs}}

\end{document}